\documentclass{article}

\usepackage[preprint]{neurips_2024}

\setcitestyle{numbers}

\usepackage{subcaption}

\usepackage[utf8]{inputenc} 
\usepackage[T1]{fontenc}    
\usepackage{hyperref}       
\usepackage{url}            
\usepackage{booktabs}       
\usepackage{amsfonts}       
\usepackage{nicefrac}       
\usepackage{microtype}      
\usepackage{xcolor}         
\usepackage[pdftex]{graphicx}
\usepackage{caption}
\usepackage{subcaption}
\usepackage[linesnumbered, ruled]{algorithm2e}
\SetAlCapSkip{1em}
\SetKwInput{KwParam}{Parameters}
\SetKwInput{KwHyper}{Hyperparameters}

\usepackage{tabularx}
\usepackage{array}

\newcolumntype{P}[1]{>{\centering\arraybackslash}p{#1}}
\newcolumntype{M}[1]{>{\centering\arraybackslash}m{#1}}

\newcommand{\archname}{Kraken}

\title{Kraken: Inherently Parallel Transformers For Efficient Multi-Device Inference}

\begin{document}

\author{
  Rohan Baskar Prabhakar\\
  Princeton University\\
  rohanbp@princeton.edu\\
  \And
  Hengrui Zhang \\
  Princeton University \\
  hengrui.zhang@princeton.edu\\
  \AND
  David Wentzlaff \\
  Princeton University \\
  wentzlaf@princeton.edu\\
}

\maketitle

\begin{abstract}

Large Transformer networks are increasingly used in settings where low inference latency can improve the end-user experience and enable new applications.
However, autoregressive inference is resource intensive and requires parallelism for efficiency.
Parallelism introduces collective communication that is both expensive and represents a phase when hardware resources are underutilized.
Towards mitigating this, \archname~is an evolution of the standard Transformer architecture that is designed to complement existing tensor parallelism schemes for efficient inference on multi-device systems.
By introducing a fixed degree of intra-layer model parallelism, the architecture allows collective operations to be overlapped with compute, decreasing latency and increasing hardware utilization.
When trained on OpenWebText, \archname~models reach a similar perplexity as standard Transformers while also preserving their language modeling capabilities when evaluated on the SuperGLUE benchmark.
Importantly, when tested on multi-GPU systems using TensorRT-LLM engines, \archname~speeds up Time To First Token by a mean of $35.6\%$ across a range of model sizes, context lengths, and degrees of tensor parallelism.

\end{abstract}

\section{Introduction}

Deep neural networks based on the Transformer architecture~\cite{original_attention} have become the de facto choice for a variety of tasks involving sequences, especially in natural language processing and computer vision~\cite{gpt2}~\cite{lms_fewshot_learners}~\cite{vit}.
Their capabilities, particularly in language modeling, have been driven by a rapid increase in parameter count~\cite{lms_fewshot_learners}.
Today's largest language models have up to a trillion parameters~\cite{switch_transformers} and consequently demand more efficiency from the systems used to train and serve them.
This has necessitated the need for many techniques and optimizations that focus on improving the performance of both algorithms and systems~\cite{multi_query}~\cite{speculative_decoding}~\cite{flash_attention}~\cite{ring_attention}~\cite{h100_trans_engine}.

Large models are often used in interactive applications where latency is an important metric that dictates the quality of the end-user experience~\cite{applications_of_lms}.
A typical web search takes about $0.2$ seconds but the Time To First Token (TTFT) for large models can be up to a few seconds (depending on context length, model size, and available hardware)~\cite{pope2023efficiently}.
Additionally, because it is not always feasible to run models on local hardware, they are served to users via datacenters that use multi-device compute nodes, adding to latency constraints.
Increasingly, language models are also used as intermediate steps in longer processes such as augmenting web searches or retrieving data from a database~\cite{applications_of_lms}.
The rising prevalence of such multi-step applications makes reducing inference latency even more critical.

Continuing this theme, this work focuses on reducing the latency cost of the collective operations introduced by tensor parallelism~\cite{megatron_lm} in the forward pass.
In particular, it introduces \archname, a variation of the standard Transformer architecture~\cite{gpt2}\cite{lms_fewshot_learners} that reduces the amount of inter-device communication and allows remaining collective operators to be overlapped with compute.
\archname~models have a fixed degree of innate model parallelism that allows computational graphs on each device to run independently without having to wait for the results of collective operators.
The architecture is designed to complement the topology of multi-device setups such as nodes in typical datacenters and DGX~\cite{dgx_h100} systems.
By designing the model architecture to account for characteristics of the hardware, our approach increases compute utilization and allows more efficient inference.

We evaluate the improvements \archname~offers over standard Transformers in two key aspects: model quality and inference latency.
For the former, we train a series of \archname~models with varying degrees of parallelism and parameter count on OpenWebText~\cite{openwebtext} and compare them with the GPT-2~\cite{gpt2} family of models on the SuperGLUE suite of benchmarks~\cite{super_glue}.
We then implement \archname~using the TensorRT-LLM library~\cite{trt_llm} and measure the Time To First Token (TTFT) given various model sizes and context lengths to illustrate the efficiency gains when collective operators are no longer on the critical path.
We find that while maintaining the language modeling capabilities of standard Transformers, \archname~models speedup the Time To First Token (TTFT) by a geomean of $35.6\%$ when tested across a range of model sizes, context lengths, and degrees of parallelism.

\section{Background}
\label{sec:background}

\subsection{\textbf{Decoder-Only Transformer models}}
\label{sub_sec:d_transformers}

\begin{figure}[!h]
\centering
\begin{minipage}[c]{0.48\linewidth}
\centering
\includegraphics[scale=0.21,trim={0 100 220 0}]{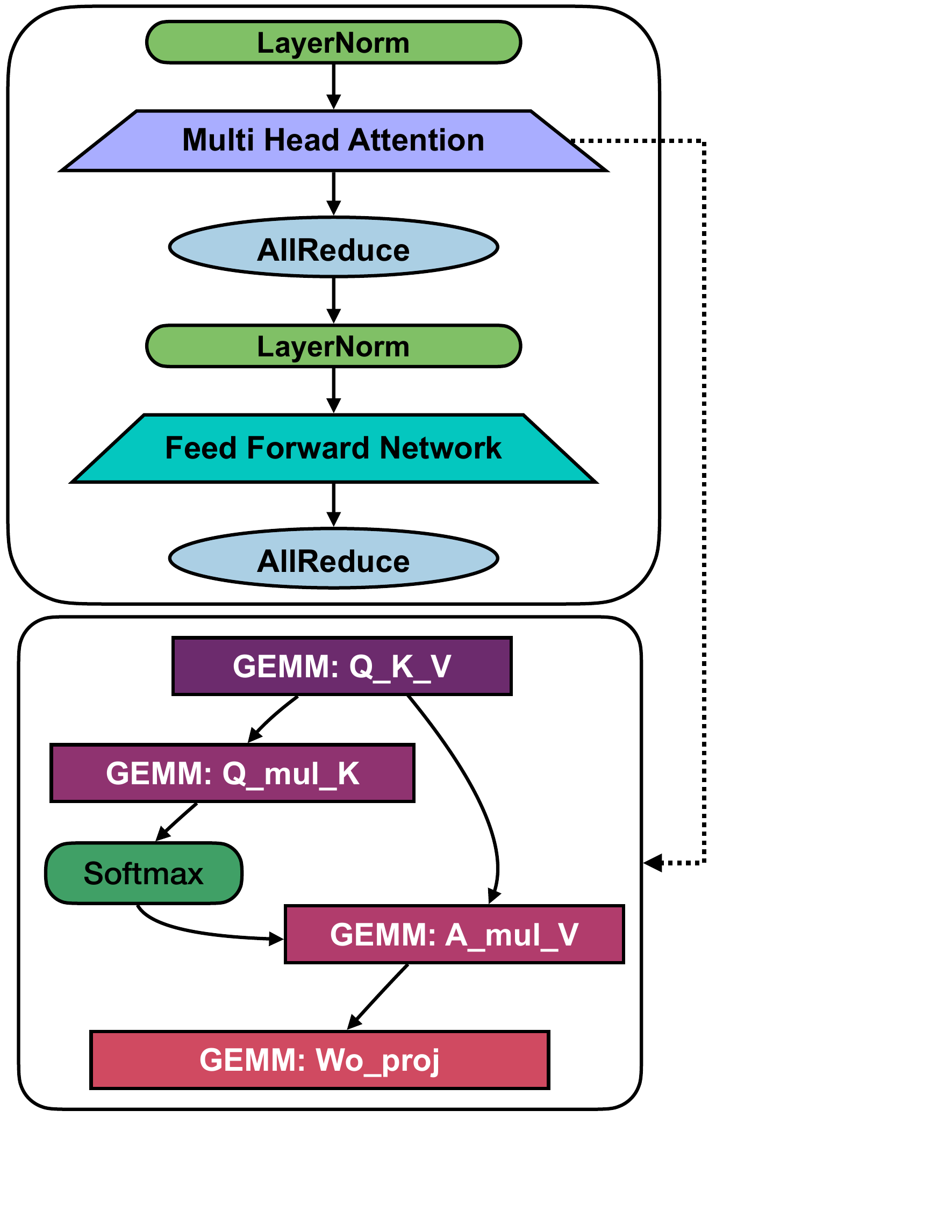}
\caption{\textbf{One layer of a standard Transformer} consisting of Multi-Head Attention (also shown) followed by a FeedForward Network.
Residual connections have been omitted.
}
\label{fig:transformer_layer}
\end{minipage} \hfill
\begin{minipage}[c]{0.48\linewidth}
\centering
\includegraphics[scale=0.44]{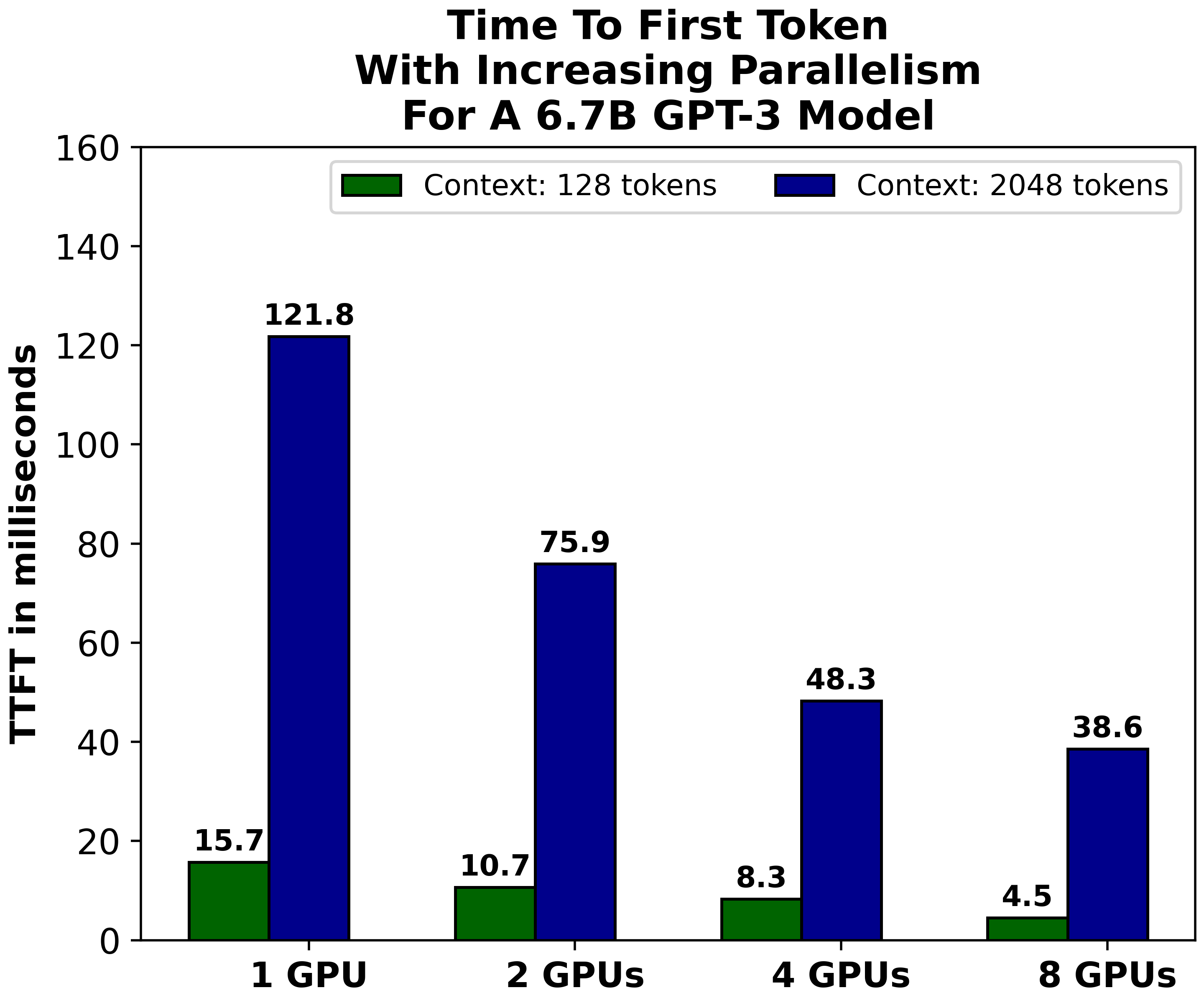}
\caption{\textbf{Increasing the degree of tensor parallelism decreases the Time To First Token.}
Even when weights and KV cache fit on device memory, parallelism can be worthwhile.
These numbers are for a 6.7B parameter GPT-3 like model and were collected using TensorRT-LLM engines on our evaluation platform: an HGX A100 40GB system.}
\label{fig:tp_benefits}
\end{minipage}
\end{figure}

We will briefly discuss the forward pass of decoder-only Transformer (DTransformer) models that use self-attention mechanisms to perform language modeling~\cite{formal_algorithms} with a focus on some commonly used inference optimizations.
Given an input sequence $x$ consisting of tokens belonging to a vocabulary $V$, such models return a probability distribution over the vocabulary that describes what the next token in $x$ could be i.e., the model is trained to estimate $P(x[\textit{l + 1}] \ |\ x[1:\textit{l}])$ where $\textit{l}$ is the initial length of $x$.
To compute the output logits in the forward pass, $x$ is converted to a sequence of embeddings that incorporate information about each token and its position in the sequence.
These embeddings are used as input to a stack of Transformer layers one of which is depicted in Figure~\ref{fig:transformer_layer}.

Such layers comprise the bulk of the compute in the model and consist of a Multi-Head Attention (MHA) block followed by a FeedForward Network (FFN or Multi-Layer Perceptron).
The FFN typically consists of two linear transformations with a non-linear activation function in between.
On the other hand, Multi-Head Attention (Figure~\ref{fig:transformer_layer}) implements scaled dot-product attention i.e., each head computes $\texttt{Softmax}(\frac{QK^{T}}{\sqrt{h}})V$ where $Q,K,V \in \mathbb{R}^{h \times l}$.
Here, $h$ is the head dimension and $l$ is the sequence length.
The General Matrix Multiply (GEMM) $Wo\_proj$ combines the outputs of the different heads.
The activations of the last layer are used as input to the unembedding operation.
Henceforth, we will refer to the GPT-2,3~\cite{gpt2}\cite{lms_fewshot_learners} like construction as the standard Transformer architecture.
This variant uses MHA, has sequential Attention and FFN blocks, expands the embedding dimension from $d$ to $4d$ in the hidden layer of the FFN, uses Gaussian Error Linear Unit (GELU) non-linearities, and places Layer Norm operators before the MHA and FFN i.e., is a Pre-LN Transformer.

A widely used optimization during inference involves caching the Key and Value matrices (from GEMM $Q\_K\_V$ in Figure~\ref{fig:transformer_layer}) of each token in a KV cache that is stored in memory.
This memoization has the effect of breaking up autoregressive inference~\cite{splitwise}\cite{llmcompass} into two distinct steps: 1) Prefill (when the first token and the KV cache are generated) and 2) Decode(for all subsequent tokens).
Both these steps have distinct runtime characteristics with Prefill being more compute bound and Decode being more memory bandwidth bound~\cite{llmcompass}.
The KV cache entry for the next generated token is appended during each Decode step.
Prefill, measured by TTFT, typically takes much longer than a single Decode step and will be the focus of this work.

\subsection{\textbf{Tensor parallelism}}
\label{sub_sec:tensor_parallelism}

Serving large models in multi-device settings requires parallelization schemes and strategies that partition model weights, activations, and incoming inference requests~\cite{megatron_lm}~\cite{orca_serving_system}~\cite{paged_attention}.
We will describe two commonly used schemes: tensor parallelism and pipeline parallelism. Given a model with $L$ layers and a node with $N$ devices, a pipeline parallel approach involves placing groups of $\left\lceil L/N \right\rceil$ layers onto each device.
A single forward pass requires moving intermediate activations through the different devices in sequence.
This approach increases the latency of an individual request but improves overall system throughput as measured in terms of the number of requests served per unit time.

In contrast, tensor parallelism involves partitioning individual computational blocks across the different devices and using collectives to aggregate/distribute results.
As shown in Figure~\ref{fig:tp_benefits}, this form of parallelism can significantly lower inference latency.
There are several possible strategies~\cite{pope2023efficiently}~\cite{overlap_comp_asplos2023} to achieve distributed tensor parallelism in Transformer models but this discussion will focus on the widely used scheme introduced by Shoeybi et al.~\cite{megatron_lm} which is well-suited for multi-GPU settings.
This scheme introduces two AllReduce operations per layer and takes advantage of the implicitly parallel nature of Multi-Head Attention with optimal partitioning of the FeedForward Network.
In each layer, contiguous groups of Attention heads are placed across different devices and the $Wo\_proj$ matrix that is used to combine the output of the different heads is partitioned across rows.
The output of the MHA block is retrieved by reducing the local output of all devices; this introduces one AllReduce as shown in Figure~\ref{fig:transformer_layer}.
Similarly, in the FFN block, the $W1\_proj$ weight matrix is partitioned across rows and the $W2\_proj$ weight matrix is partitioned across columns.
The output of the FFN is computed by reducing the local output of all devices, thereby introducing another AllReduce operation.
The focus of this work will be on reducing and hiding the runtime impact of these collectives introduced by tensor parallelism.

\subsection{\textbf{Multi-Device systems}}
\label{sub_sec:multi_device_systems}

Given the extensive amount of compute and memory capacity required to efficiently serve large models, most widely used systems are node-based configurations where each node has a small number (between $4$ and $16$) of devices.
These devices typically take the form of Graphics Processing Units (GPUs) or Tensor Processing Units (TPUs).
The discussion in this work will focus on GPUs but we expect that our findings will also be of relevance to other choices of accelerators such as TPUs.
A typical device adopts a wide, Single Instruction, Multiple Data (SIMD) architecture and consists of several compute cores/blocks along with on-chip, volatile memory and off-chip Dyanmic Random Access Memory (DRAM) or High Bandwidth Memory (HBM).
Accessing data at levels of the system physically closer to compute cores such as scratchpad memory or caches is typically much faster and more efficient compared to accessing HBM/DRAM.
Techniques like tiled matrix multiplication~\cite{tiled_gemm} and FlashAttention~\cite{flash_attention} account for this characteristic, considerably speeding up implementations.

Devices within a node are configured in a topology and linked by interconnects such as Peripheral Component Interconnect Express (PCIe), NVLink, and NVSwitch.
The different standards balance versatility and cost with performance.
For example, a topology that uses PCIe switches across some connections will have less overall bandwidth than a system that uses solely NVLink/NVSwitch.
Inter-device communication primitives are provided to other software by libraries like NCCL~\cite{nccl} and RCCL~\cite{rccl}.
Communication is comparatively expensive and represents a phase in the forward pass where compute cores are mostly idle.
This work strives to extend the IO-aware approach used by techniques like FlashAttention~\cite{flash_attention} towards the multi-device setting.

\section{\textbf{\archname: Model architecture}}
\label{sec:architecture}

\begin{figure}
\includegraphics[width=\textwidth,trim={0 2cm 0 0}]{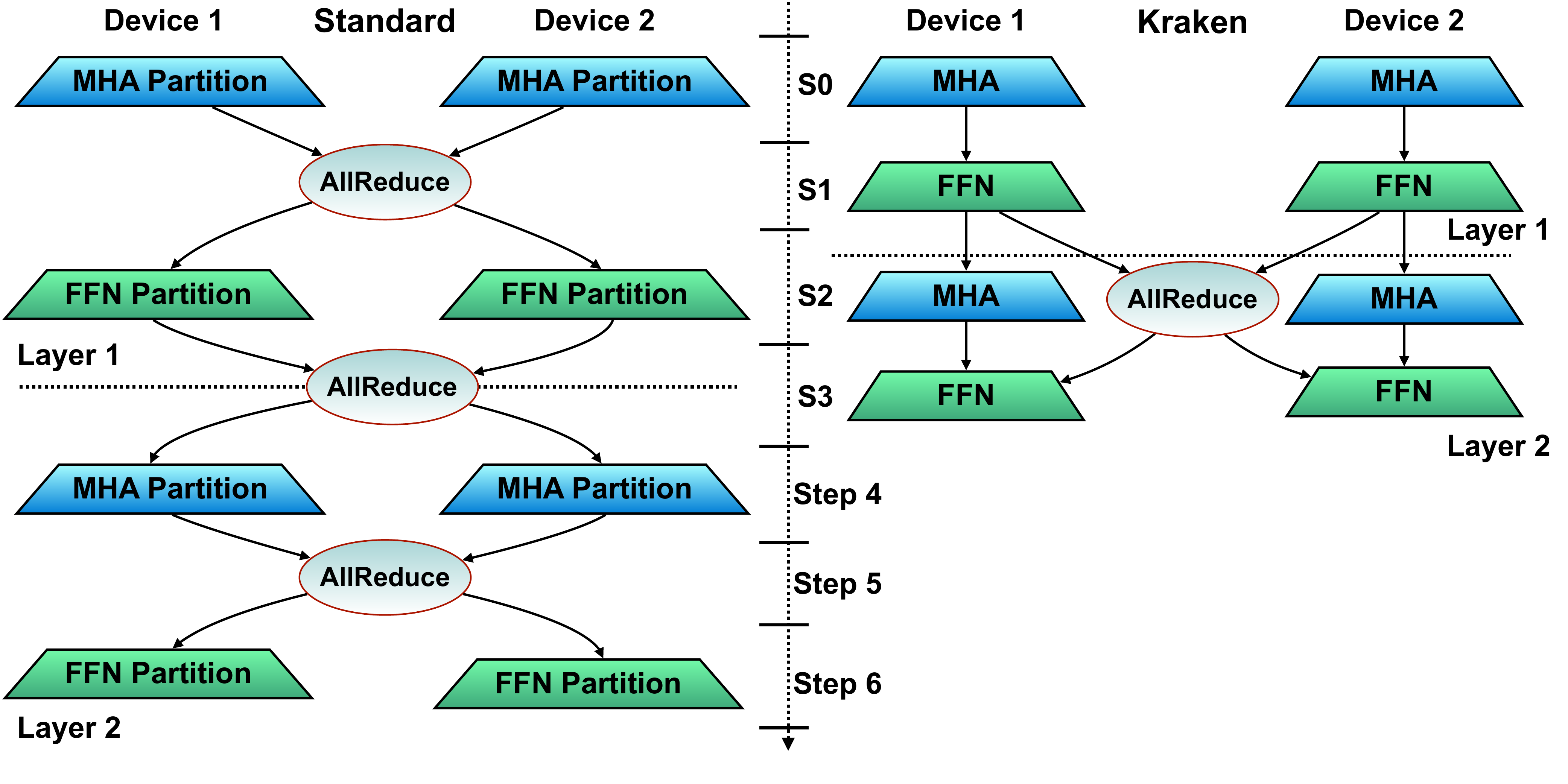}
\caption{\textbf{Parallelizing two standard Transformer layers compared to executing two layers of a \archname~Transformer with 2-way parallelism}. \archname~Transformers have fewer AllReduce ops and these can be run concurrently with the Multi-Head Attention in the next layer. Step lengths are illustrative and not indicative of how much wall-clock time a particular operation might actually require.}
\label{fig:kraken_layer}
\vspace{-0.5cm}
\end{figure}

\subsection{\textbf{Layer construction}}
\label{sub_sec:layer_construction}

The primary objective of \archname~is to preserve as much of the standard Transformer (GPT-2,3 like)\cite{gpt2}\cite{lms_fewshot_learners} architecture as possible while reducing the latency cost of the collective communication introduced by tensor parallelism.
To achieve this, we allow each of the individual shards of a parallelized Transformer layer to behave as independent, smaller layers.
More precisely, as depicted in Figure~\ref{fig:kraken_layer}, instead of sharding the Multi-Head Attention and FeedForward Network blocks, each MHA and FFN block is replaced with a smaller, independent block.
This introduces \textit{a fixed degree of parallelism} that is chosen at the outset of training and takes into consideration the most common hardware deployment target.
For example, if a model will be mostly served on nodes with eight GPUs each, a reasonable choice would be to use either $4$-way or $8$-way parallelism (depending on the size of the model).
The former would be suitable for smaller models, allowing each node to serve two different inference requests at the same time.

\begin{algorithm}
\SetKwInput{KwData}{Input}
\SetKwInput{KwResult}{Output}
\newcommand\commentstyle[1]{\footnotesize\ttfamily\textcolor{blue}{#1}}
\SetCommentSty{commentstyle}

\KwData{$x \in \mathbb{R}^{l \times d}$}
\DontPrintSemicolon
\KwResult{$y \in  \mathbb{R}^{l \times d}$}
$ residual = x$\;
\tcc{The first layer replaces the AllReduce with the identity operator}
$ y = \texttt{AllReduce}(x)$\;
$ x = \texttt{LayerNorm}(x)$\;
$ x =  residual + \texttt{MultiHeadAttention}(x)$\;
$ residual = x$\;
\tcc{The output of the \texttt{AllReduce} is used only here allowing it to be overlapped with \texttt{Attention}}
$ x = \texttt{LayerNorm}(x + y)$\;
$y =  residual + \texttt{FeedForwardNetwork}(x)$\;
return $y$\;
\caption{\archname~ Sub-Layer: Forward Pass}
\label{alg:forward_pass}
\end{algorithm}

Consequently, the only inter-device dependency is an AllReduce operation at the end of every layer.
This collective represents the only interchange of activations between groups of sub-layers.
Its output is used in the FFN block of the next layer and not in the MHA block.
As shown in Figure~\ref{fig:kraken_layer}, this allows for overlapping the compute in the MHA block with the AllReduce, effectively removing most inter-device communication from the critical path.
Much like how positional embeddings are added to the token embeddings prior to the first layer~\cite{formal_algorithms}, we chose to use element-wise addition to combine the outputs of the various sub-layers from the previous layer.
This occurs prior to the Layer Norm before the FFN.
The forward pass of each sub-layer, including residual connections, is also described in Algorithm~\ref{alg:forward_pass}.
All pretrained models in our evaluation use the GELU activation function in the FFN.
In initial experiments that explored different constructions, we scaled the weights of the residual layer by $\frac{1}{\sqrt{L*N}}$ using a similar line of reasoning as that used by Radford et al.~\cite{gpt2}.
Here, $L$ is the number of layers, and $N$ is the degree of parallelism.
We did not remove this initialization scheme in subsequent experiments even though accumulation along residual connections was limited to groups of $L$ sub-layers.

The token and positional embeddings are shared across all sub-layers of the first layer i.e., there is still one set of embeddings to maintain compatibility with weight tying~\cite{inan_tying}~\cite{press2016_tying}.
After the last layer, we combine the outputs of the different sub-layers using a linear transformation with weights $W_{concat} \in \mathbb{R}^{d*N \times d}$ where $d$ is the embedding dimension and $N$ is the degree of parallelism.
The output of this transformation is used as input to the unembedding.
Depending on the implementation, this linear layer introduces the only blocking collective in the computational graph.

\subsection{\textbf{Deriving model configurations for a fixed parameter budget}}
\label{sub_sub_sec:deriving_model_configs}

Increasing the degree of parallelism while keeping other hyperparameters like the embedding dimension constant will increase the parameter count.
Instead, using a configuration of a standard Transformer as the basis, we make the following two modifications to derive a \archname~configuration that has the same number of parameters:
\begin{itemize}
    \item First, the hidden state expansion in the FFN is reduced from $4d$ to $2d$ where $d$ is the embedding/model dimension.
    \item Given the number of parameters $P$, degree of parallelism $N$, number of layers $L$, and vocabulary size $V$, we derive a closed form expression for $P$ and solve for $d$ i.e., $P = V*d + \Sigma_{i=1}^{L}N*((3*(d \times d) + (d \times d)) + 2*(d \times 2d)) $ where the term  $((3*(d \times d) + (d \times d))$ comes from MHA and $2*(d \times 2d))$ comes from the FFN.
\end{itemize}

\section{\textbf{Evaluation}}
\label{sec:evaluation}

\subsection{Model configurations and perplexity}
\label{sub_sec:model_confs_perplexity}

\begin{table}
\centering
\caption{\textbf{Model configurations and perplexity on OpenWebText for \archname~models compared to similarly sized GPT-2 models.} Lower perplexity is better.}
\resizebox{\textwidth}{!}{
\begin{tabular}{ P{0.225\textwidth}cP{0.125\textwidth}P{0.1\textwidth}P{0.1\textwidth}P{0.125\textwidth}}
\toprule
\textbf{Model} & \textbf{Layers} & \textbf{Embedding Dimension} &  \textbf{Attention Heads} & \textbf{Total Params.} & \textbf{Validation Perplexity} \\
\toprule
  GPT-2 & 12 & 768 & 12 & 117M & 20.64 \\ 
  \archname\ 2-way & 12 & 678 & 12 & 124M & 18.89  \\
 \archname\ 4-way & 12 & 504 & 12 & 124.5M & 18.56 \\
  \archname\ 6-way & 12 & 418 & 12 & 123.2M & 19.22 \\
\midrule
  GPT-2 Medium &  24 & 1024 & 16 & 345M & 14.87 \\
  \archname\ 2-way &  24 & 888 & 16 & 350M & 14.40  \\
  \archname\ 4-way &  24 & 644 & 16 & 353.4M & 14.71  \\
 \midrule
 GPT-2 Large & 24 & 1280 & 20 & 762M & 13.69 \\
 \archname\ 4-way & 24 & 960 & 16 & 761M & 13.09 \\
\bottomrule
\end{tabular}}
\label{tab:model_configurations}
\end{table}

To evaluate the language modeling performance of \archname, we train a series of models up to 760 million parameters large and with varying degrees of parallelism on OpenWebText~\cite{openwebtext}.
This allows us to compare the performance of the architecture with the GPT-2~\cite{gpt2} family of models.
Because of limited access to compute, we do not exhaustively search for hyperparameters and stop training at $150$ billion tokens in contrast to the about $300$ billion tokens that language models of such sizes are typically trained for~\cite{gpt2}\cite{mamba}.
Table~\ref{tab:model_configurations} details the embedding dimensions, number of layers, and other hyperparameters for each configuration.
For \archname~configurations, the number of Attention heads in each layer is summed across all sub-layers.
The context length was set at $1024$ tokens.
Models similar in size were trained using the same learning rate schedule and for the same number of gradient steps.
More details about the training setup, required compute, and how each configuration was derived can be found in Appendix~\ref{appendix:training_setup_compute}.

Perplexity measurements for GPT-2 models are provided to add context to the results with the caveat that all \archname~models were trained on OpenWebText but GPT-2 models were trained on the closed-source WebText dataset.
We would expect that all things equal, \archname~models will have lower perplexity unless the GPT-2 models were subsequently fine-tuned on OpenWebText.
Nonetheless, even when trained over a smaller number of tokens, \archname~models reach a similarly low perplexity as standard Transformers.

\subsection{\textbf{Performance On SuperGLUE}}
\label{sub_sec:superglue_perf}

\begin{table}
\caption{\textbf{Zero-Shot Performance on SuperGLUE.} ReCoRD uses the F1 score as the evaluation metric. All other benchmarks use accuracy.}
\centering
\resizebox{\textwidth}{!}{
\begin{tabular}{ P{0.15\textwidth}ccccccccc } 
\toprule
\textbf{Model} & \textbf{BoolQ} & \textbf{RTE} & \textbf{CB} & \textbf{COPA} & \textbf{ReCoRD} & \textbf{WIC} & \textbf{WSC} & \textbf{MultiRC} & \textbf{Average}\\ 
 
\toprule
GPT-2 & 48.38 & 51.99 & 41.07 & 62.0 & 71.07 & 49.53 & 43.27 & 53.47 & 52.6 \\ 
 
\archname\ 124M (2-way) & 53.85 & 54.15 & 44.64 & 68.0 & 72.42 & 49.53 & 36.54 & 56.89 & 54.5 \\ 
 
\archname\ 124M (4-way) & 50.15 & 53.79 & 41.07 & 67.0 & 71.47 & 50.31 & 37.5 & 57.16 & 53.56 \\ 
 
\archname\ 124M (6-way) & 47.92 & 56.68 & 8.93 & 69.0 & 70.69 & 50.16 & 53.85 & 53.82 & 51.38 \\ 
 
\midrule
GPT-2 Medium & 58.53 & 53.07 & 42.86 & 68.0 & 79.43 & 50.0 & 41.35 & 52.58 & 55.73 \\ 
 
\archname\ 355M (2-way) & 60.06 & 51.62 & 41.07 & 69.0 & 80.04 & 50.0 & 35.58 & 56.91 & 55.54 \\ 
 
\archname\ 355M (4-way) & 61.68 & 55.23 & 35.71 & 72.0 & 79.01 & 50.0 & 36.54 & 57.2 & 55.92 \\ 
 
\midrule
GPT-2 Large & 60.55 & 52.71 & 41.07 & 72.0 & 81.95 & 49.69 & 44.23 & 48.56 & 56.34 \\ 
 
\archname\ 760M (4-way) & 60.58 & 49.1 & 10.71 & 73.0 & 82.04 & 50.0 & 36.54 & 51.65 & 51.7 \\ 
 
\bottomrule

\end{tabular}}
\label{tab:superglue_zero_shot}
\end{table}

\begin{table}
\centering
\caption{\textbf{Three-Shot Performance on SuperGLUE.} ReCoRD uses the F1 score as the evaluation metric. All other benchmarks use accuracy.}
\resizebox{\textwidth}{!}{
\begin{tabular}{ P{0.15\textwidth}ccccccccc } 
\toprule
\textbf{Model} & \textbf{BoolQ} & \textbf{RTE} & \textbf{CB} & \textbf{COPA} & \textbf{ReCoRD} & \textbf{WIC} & \textbf{WSC} & \textbf{MultiRC} & \textbf{Average}\\ 

\toprule
GPT-2 & 53.76 & 47.65 & 42.86 & 60.0 & 70.12 & 50.0 & 51.92 & 50.62 & 53.37 \\ 
 
\archname\ 124M (2-way) & 55.14 & 48.74 & 44.64 & 61.0 & 68.83 & 49.69 & 47.12 & 50.56 & 53.22 \\ 
 
\archname\ 124M (4-way) & 59.05 & 44.04 & 48.21 & 64.0 & 69.94 & 47.49 & 51.92 & 54.17 & 54.85 \\ 
 
\archname\ 124M (6-way) & 54.28 & 46.93 & 42.86 & 63.0 & 69.17 & 49.37 & 44.23 & 51.96 & 52.72 \\ 
 
\midrule
GPT-2 Medium & 60.58 & 47.65 & 44.64 & 68.0 & 78.35 & 47.65 & 50.0 & 53.22 & 56.26 \\ 
 
\archname\ 355M (2-way) & 50.73 & 53.07 & 37.5 & 68.0 & 79.01 & 49.06 & 66.35 & 51.34 & 56.88 \\ 
 
\archname\ 355M (4-way) & 61.8 & 51.99 & 50.0 & 73.0 & 77.97 & 47.49 & 44.23 & 54.17 & 57.58 \\ 
 
\midrule
GPT-2 Large & 60.58 & 51.99 & 39.29 & 70.0 & 81.05 & 47.81 & 60.58 & 51.53 & 57.85 \\ 
 
\archname\ 760M (4-way) & 55.75 & 53.79 & 42.86 & 68.0 & 81.03 & 45.77 & 62.5 & 53.47 & 57.9 \\ 
 
\bottomrule

\end{tabular}}
\label{tab:superglue_three_shot}
\end{table}

We used the SuperGLUE benchmark suite to evaluate performance on language tasks~\cite{super_glue}.
All benchmarks were scored on accuracy except for ReCoRD which uses the F1 Score instead.
No finetuning or training was performed for any combination of model and benchmark.
Table~\ref{tab:superglue_zero_shot} contains results for Zero-shot performance and Table~\ref{tab:superglue_three_shot} presents performance in the Three-shot setting.
Scores were calculated using the Language Model Evaluation Harness~\cite{lm_eval_harness} with the default choice of prompts and scoring metrics for option \texttt{lm-eval-SuperGLUE v1}.
As conveyed by these results, \archname~largely preserves the language modeling capabilities of the standard Transformer architecture.
We expect that the gap between standard Transformers on language tasks will close further if we train the models for longer.

\subsection{\textbf{Evaluation platform}}
\label{sub_sec:evaluation_platform}

To measure improvements in inference latency, we used the TensorRT-LLM~\cite{trt_llm} library to create engines and compare \archname~models with other widely used dense model architectures.
The library provides an interface to define popular Transformer models and bundles a collection of kernels, plugins, and other optimizations that can be used to create efficient TensorRT engines (containing model weights) and serve them on systems with GPUs.
All experiments were conducted on a 8 x A100 GPU machine with NVSwitch and 40GB of HBM memory per GPU.

\subsection{\textbf{Speedup in Time To First Token}}
\label{sub_sec:first_token_latency}

\begin{table}
\centering
\caption{\textbf{Configurations for the different model engines used to compare TTFT.} For models of similar sizes, hyperparameters are shared for the GPT-like and Parallel Attention + FeedForward variants. 4-way denotes \archname~configurations used when evaluating tensor parallelism across 4 devices and likewise for 8-way.}
\resizebox{\textwidth}{!}{
\begin{tabular}{ P{0.06\textwidth}P{0.06\textwidth}P{0.06\textwidth}P{0.1\textwidth}P{0.1\textwidth}|P{0.08\textwidth}P{0.1\textwidth}P{0.08\textwidth}|P{0.1\textwidth}P{0.1\textwidth}P{0.08\textwidth}} 
\toprule
\textbf{Model Size} & \textbf{Layers} & \textbf{dModel} & \textbf{Params. Per Layer} & \textbf{Attention Heads} & \textbf{\archname\ 4-way dModel} & \textbf{4-way Params. Per Layer} & \textbf{4-way Attention Heads} & \textbf{\archname\ 8-way dModel} & \textbf{8-way Params. Per Layer} & \textbf{8-way Attention Heads} \\ 
\toprule
1.3B & 24 & 2048 & 50.3M & 16 & 1248 & 49.9M & 12 & 960 & 59.0M & 10  \\
6.7B & 32 & 4096 & 201.3M & 32 & 2496 & 199.4M & 24  & 1920 & 235.9M & 20 \\
13B & 40 & 5140 & 317.0M & 40 & 3120 & 311.5M & 30 & 2304 & 339.7M & 32 \\
65B & 80 & 8192 & 805.3M & 64 & 4992 & 797.4M & 39 & 3648 & 851.7M & 38 \\
175B & 96 & 12288 & 1.81B & 96 & 7424 & 1.76B & 58 & 5472 & 1.92B & 57 \\
\bottomrule
\end{tabular}}
\label{tab:engine_model_dims}
\end{table}

For comparisons with other model architectures, we build engines for standard, GPT-like configurations and GPT-J~\cite{gpt-j} like configurations as detailed in Table~\ref{tab:engine_model_dims}.
The latter serves to contrast our approach with one that runs the FFN in parallel with the Attention~\cite{palm}~\cite{gpt-j}.
Such parallel layers require only one AllReduce in a layer but unlike \archname, the collective cannot be overlapped with compute.
For similarly sized models, the only difference is the embedding dimension for the \archname~models.
All configurations follow the convention where \texttt{dModel} is divided by the number of Attention heads to calculate the size of each head.
Other configuration parameters such as the number of layers, maximum context length, and vocabulary size were the same.
The Attention heads are per sub-layer for \archname~models and parameter counts do not include biases and layer normalization.

There are two sets of engines, one for each of the three model architectures with tensor parallelism set to four and the other for 8-way parallelism.
We also did not follow the earlier convention~\ref{sub_sub_sec:deriving_model_configs} of solving for the embedding dimension of a \archname~model precisely.
This is because the available optimizations in TensorRT-LLM are compatible only with specific Attention head dimensions.
To account for this, we handpicked embedding dimensions that given an equivalent GPT-like configuration have about the same number of parameters while still being compatible with the available kernels.
This was necessary for a fair evaluation but in practice, the chosen embedding dimension should also account for the performance of the computational kernels it would map to.
Anthony et al.~\cite{choosing_efficient_dimensions} discuss this, showing how model hyperparameters can affect GEMM performance and consequently training efficiency.

\begin{figure}
\centering
\includegraphics[width=1.0\textwidth]{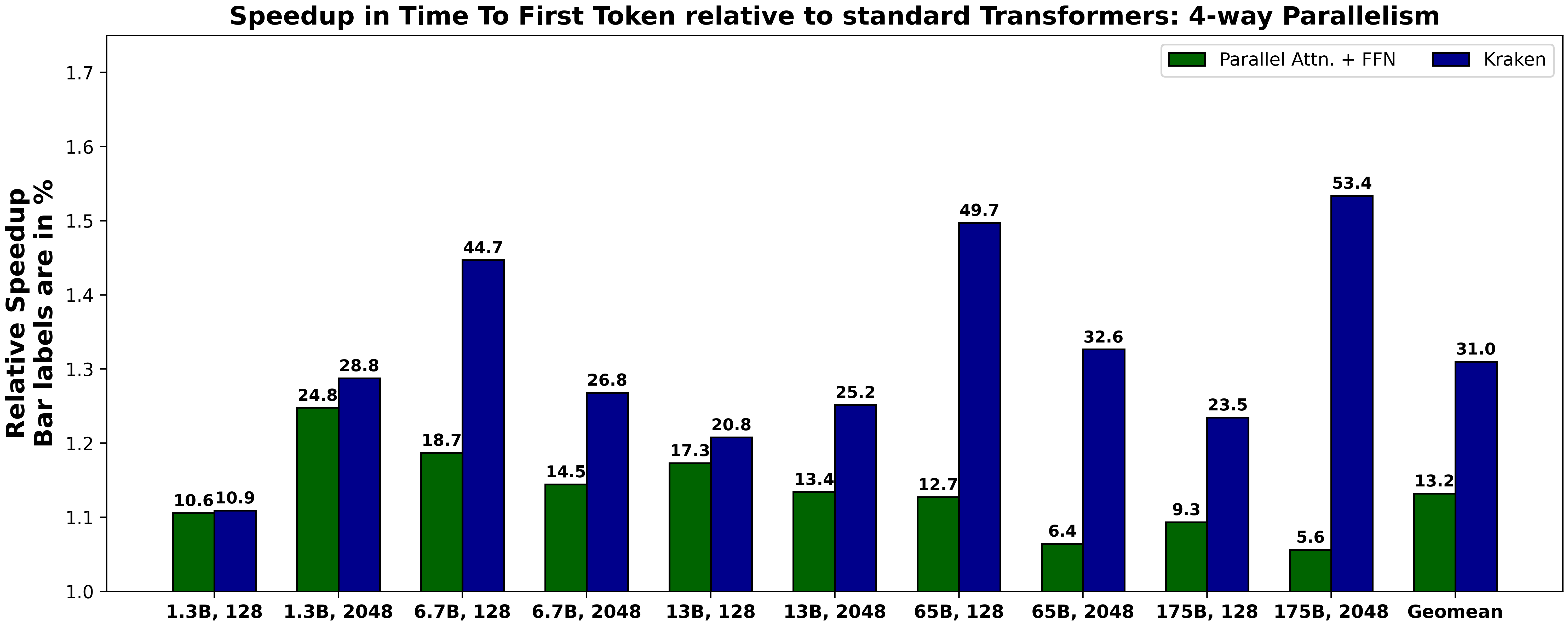}
\caption{\textbf{Speedup in Time To First Token over standard Transformers on a system that uses NVSwitch and with 4-way parallelism}. x-axis labels denote the size of the model followed by the context length. Bar labels are in percentage.}
\label{fig:latency_improvement_4}
\vspace{-0.2cm}
\end{figure}

\begin{figure}
\centering
\includegraphics[width=1.0\textwidth]{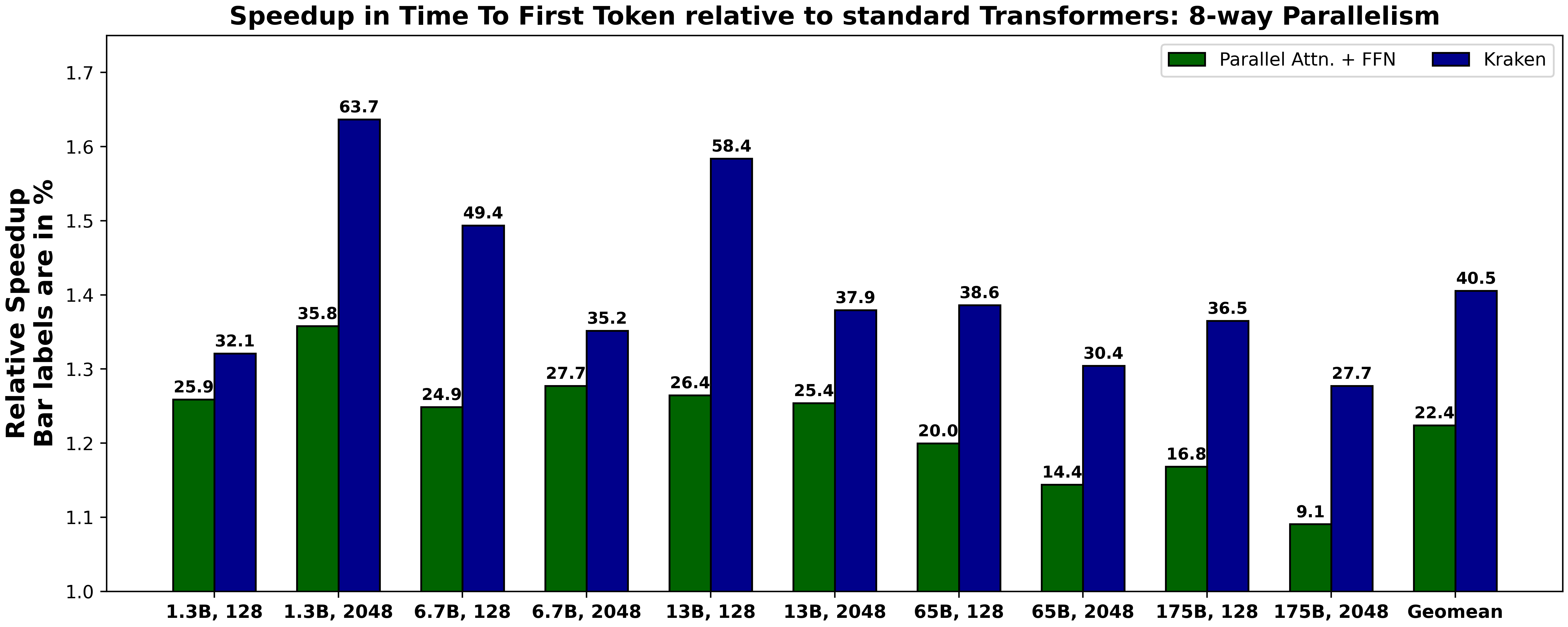}
\caption{\textbf{Speedup in Time To First Token over standard Transformers on a system that uses NVSwitch and with 8-way parallelism}. x-axis labels denote the size of the model followed by the context length. Bar labels are in percentage.}
\label{fig:latency_improvement_8}
\vspace{-0.2cm}
\end{figure}

Figure~\ref{fig:latency_improvement_4} shows the speedup in latency compared to GPT-like models on our evaluation platform and with 4-way parallelism while Figure~\ref{fig:latency_improvement_8} depicts improvements for 8-way parallelism.
Across a range of model sizes and context lengths of $128$ and $2048$ tokens, using \archname~models can improve inference latency by anywhere from $10.9\%$ to $63.7\%$.
These results are end-to-end and include the extra fully-connected layer required by \archname~models.
All results are normalized to the equivalent latency for a similar GPT-3 like model and engines were generated with random weights and \textit{fp16} precision.
Latency in terms of milliseconds is reported in Appendix~\ref{appendix:ttft_seconds}.
Importantly, our evaluation platform uses NVSwitch and consequently has considerable inter-device bandwidth ($600$ GB/s) which means communication can be relatively inexpensive.
We expect that these gains will be more pronounced on systems with less performant interconnects.

A more detailed evaluation on precisely which operators contribute to overall runtime can be found in Appendix~\ref{appendix:runtime_characterization}.
Since \archname~models replicate the Attention block by the degree of tensor parallelism, the size of the KV cache will also be larger than that of an equivalent Transformer model.
A discussion on how this affects per layer memory consumption is presented in Appendix~\ref{appendix:kv_cache_analysis}.
One way to mitigate this effect would be to replace Multi-Head Attention with either Multi-Query Attention~\cite{multi_query} or Grouped Query Attention~\cite{ainslie2023gqa}.
Additionally, in order to take advantage of the overlapped communication and compute that \archname~allows, we added a plugin to TensorRT-LLM that is described in Appendix~\ref{appendix:overlap_plugin_implementation}.

Each configuration was evaluated using the \textit{gptSessionBenchmark} that is available as part of TensorRT-LLM.
We enabled the default set of optimizations which include GPT Attention plugin, Remove Input Padding, and GEMM plugin.
Experiments were run with a batch size of $1$ but engines were built with a maximum batch size of $4$, vocabulary size of $51,200$, maximum input context length of $2048$, and maximum output length of $4096$.
For the 65B and 175B configurations, we ran into CUDA Out-of-Memory errors when running full sized engines because each device has only 40GB of memory.
To avoid this, we reduce the number of layers to $\frac{1}{4}$ the original number.
Nonetheless, the relative latency comparison should be unaffected because the runtimes of individual layers are identical.

\section{\textbf{Discussion and Limitations}}
\label{sec:discussion}

One downside to our approach is the need for expensive pretraining.
Consequently, developing techniques to distill learned weights from existing models possibly as part of the weight initialization scheme is a promising area of future work.
We also do not compare with more state-of-the-art Transformer training recipes or alternatives because of limited access to GPU compute.
Training larger \archname~models for longer on a few different datasets will permit evaluations on newer, more complex language modeling benchmarks.
The fixed degree of model parallelism also places restrictions on the optimal choice of hardware to run a model on.
For example, if we tried to deploy a model with 4-way parallelism on a system with two or six devices, we would either need to run groups of two sub-layers together or introduce more collectives.
Either approach might negate most of the latency gains offered in the first place.
The memory footprint of each layer also increases compared to standard Transformers (Appendix~\ref{appendix:kv_cache_analysis}) but this effect can be mitigated using Multi-Query~\cite{multi_query} or Grouped-Query Attention~\cite{ainslie2023gqa}.

By virtue of its architecture, \archname~evokes a comparison to Mixture-of-Experts (MoE) architectures such as the SwitchTransformer~\cite{switch_transformers}\cite{shazeer2017outrageously}.
Layers in current MoE models have a single Attention block and use a learned Router to direct each token to one of a set of FeedForward Networks.
Because the forward pass activates only a fixed fraction of the parameters, MoE models can be much larger than dense Transformers while maintaining the computational profile of inference.
Nonetheless, serving them efficiently in multi-device settings is a challenge: MoEs require dynamic routing and suffer from load balancing issues.
For instance, Huang et al.~\cite{huang2023moe} find that the required All-to-All collectives can comprise a significant fraction of inference latency.
An interesting direction of future work would be to incorporate an inter-device IO-aware approach in the construction of MoE models.

Many improvements to the standard Transformer architecture are also applicable to \archname~models.
For instance, the Attention block~\ref{alg:forward_pass} can be modified to support Multi-Query Attention~\cite{multi_query} or Grouped Query Attention~\cite{ainslie2023gqa}.
The architecture is also compatible with "drop-in" replacements for Attention that are more efficient because of time-complexity and/or sparsity~\cite{kitaev2020reformer}~\cite{choromanski2022rethinking}.
Furthermore, we expect that the notion of parallelizing individual layers of the model will also prove useful in large deep learning models that use constructions other than Attention or MLPs.
Incorporating a communication-aware approach to Neural Architecture Search~\cite{nas_transformers_survey} for Transformers is another promising area of future work.

We also expect that \archname~will be readily compatible with many existing system-aware techniques used in deploying standard Transformers.
For example, it provides an extra degree of freedom to the various partitioning strategies proposed by Pope et al.~\cite{pope2023efficiently}.
It is also compatible with techniques like FlashAttention~\cite{flash_attention}, Speculative Decoding~\cite{speculative_decoding}, fusing Attention with the FFN~\cite{blockwise_parallel}, and PagedAttention~\cite{paged_attention}.

\section{Acknowledgements}

This material is based on research sponsored by the Air Force Research Laboratory (AFRL) and the Defense Advanced Research Projects Agency (DARPA) under agreement No. FA8650-18-2-7862. The U.S. Government is authorized to reproduce and distribute reprints for Governmental purposes notwithstanding any copyright notation thereon.
The views and conclusions contained herein are those of the authors and should not be interpreted as necessarily representing the official policies or endorsements, either expressed or implied, of the Air Force Research Laboratory (AFRL), the Defense Advanced Research Projects Agency (DARPA), or the U.S. Government.
We thank Princeton Research Computing for their assistance in setting up and maintaining the necessary compute infrastructure.
We thank Ashwinee Panda for feedback and aid in brainstorming the title.
We also thank Jens Tuyls for the helpful discussions.

\bibliographystyle{plain}
\bibliography{references}

\begin{thebibliography}{10}

\bibitem{lm_eval_harness}
Eleuther AI.
\newblock Language model evaluation harness, 2022.

\bibitem{ainslie2023gqa}
Joshua Ainslie, James Lee-Thorp, Michiel de~Jong, Yury Zemlyanskiy, Federico Lebrón, and Sumit Sanghai.
\newblock Gqa: Training generalized multi-query transformer models from multi-head checkpoints, 2023.

\bibitem{choosing_efficient_dimensions}
Quentin Anthony, Jacob Hatef, Deepak Narayanan, Stella Biderman, Stas Bekman, Junqi Yin, Aamir Shafi, Hari Subramoni, and Dhabaleswar Panda.
\newblock The case for co-designing model architectures with hardware, 2024.

\bibitem{bar2006second}
Roy Bar~Haim, Ido Dagan, Bill Dolan, Lisa Ferro, Danilo Giampiccolo, Bernardo Magnini, and Idan Szpektor.
\newblock The second {PASCAL} recognising textual entailment challenge.
\newblock In {\em Proceedings of the Second {PASCAL} Challenges Workshop on Recognising Textual Entailment}, 2006.

\bibitem{bentivogli2009fifth}
Luisa Bentivogli, Ido Dagan, Hoa~Trang Dang, Danilo Giampiccolo, and Bernardo Magnini.
\newblock The fifth {PASCAL} recognizing textual entailment challenge.
\newblock 2009.

\bibitem{lms_fewshot_learners}
Tom~B. Brown, Benjamin Mann, Nick Ryder, Melanie Subbiah, Jared Kaplan, Prafulla Dhariwal, Arvind Neelakantan, Pranav Shyam, Girish Sastry, Amanda Askell, Sandhini Agarwal, Ariel Herbert-Voss, Gretchen Krueger, Tom Henighan, Rewon Child, Aditya Ramesh, Daniel~M. Ziegler, Jeffrey Wu, Clemens Winter, Christopher Hesse, Mark Chen, Eric Sigler, Mateusz Litwin, Scott Gray, Benjamin Chess, Jack Clark, Christopher Berner, Sam McCandlish, Alec Radford, Ilya Sutskever, and Dario Amodei.
\newblock Language models are few-shot learners, 2020.

\bibitem{nas_transformers_survey}
Krishna~Teja Chitty-Venkata, Murali Emani, Venkatram Vishwanath, and Arun~K. Somani.
\newblock Neural architecture search for transformers: A survey.
\newblock {\em IEEE Access}, 10:108374--108412, 2022.

\bibitem{choromanski2022rethinking}
Krzysztof Choromanski, Valerii Likhosherstov, David Dohan, Xingyou Song, Andreea Gane, Tamas Sarlos, Peter Hawkins, Jared Davis, Afroz Mohiuddin, Lukasz Kaiser, David Belanger, Lucy Colwell, and Adrian Weller.
\newblock Rethinking attention with performers, 2022.

\bibitem{palm}
Aakanksha Chowdhery, Sharan Narang, Jacob Devlin, Maarten Bosma, Gaurav Mishra, Adam Roberts, Paul Barham, Hyung~Won Chung, Charles Sutton, Sebastian Gehrmann, Parker Schuh, Kensen Shi, Sasha Tsvyashchenko, Joshua Maynez, Abhishek Rao, Parker Barnes, Yi~Tay, Noam Shazeer, Vinodkumar Prabhakaran, Emily Reif, Nan Du, Ben Hutchinson, Reiner Pope, James Bradbury, Jacob Austin, Michael Isard, Guy Gur-Ari, Pengcheng Yin, Toju Duke, Anselm Levskaya, Sanjay Ghemawat, Sunipa Dev, Henryk Michalewski, Xavier Garcia, Vedant Misra, Kevin Robinson, Liam Fedus, Denny Zhou, Daphne Ippolito, David Luan, Hyeontaek Lim, Barret Zoph, Alexander Spiridonov, Ryan Sepassi, David Dohan, Shivani Agrawal, Mark Omernick, Andrew~M. Dai, Thanumalayan~Sankaranarayana Pillai, Marie Pellat, Aitor Lewkowycz, Erica Moreira, Rewon Child, Oleksandr Polozov, Katherine Lee, Zongwei Zhou, Xuezhi Wang, Brennan Saeta, Mark Diaz, Orhan Firat, Michele Catasta, Jason Wei, Kathy Meier-Hellstern, Douglas Eck, Jeff Dean, Slav Petrov, and Noah Fiedel.
\newblock Palm: Scaling language modeling with pathways, 2022.

\bibitem{rccl}
AMD Corp.
\newblock Rocm collective communication library, 2024.

\bibitem{h100_trans_engine}
NVIDIA Corp.
\newblock H100 transformer engine supercharges ai training, 2022.

\bibitem{dgx_h100}
NVIDIA Corp.
\newblock Nvidia dgx h100, 2023.

\bibitem{nccl}
NVIDIA Corp.
\newblock Nvidia collective communications library, 2024.

\bibitem{trt_llm}
NVIDIA Corp.
\newblock Tensorrt-llm, 2024.

\bibitem{dagan2006pascal}
Ido Dagan, Oren Glickman, and Bernardo Magnini.
\newblock The {PASCAL} recognising textual entailment challenge.
\newblock In {\em Machine learning challenges. evaluating predictive uncertainty, visual object classification, and recognising tectual entailment}, pages 177--190. Springer, 2006.

\bibitem{flash_attention}
Tri Dao, Daniel~Y. Fu, Stefano Ermon, Atri Rudra, and Christopher Ré.
\newblock Flashattention: Fast and memory-efficient exact attention with io-awareness, 2022.

\bibitem{demarneffe:cb}
Marie-Catherine De~Marneffe, Mandy Simons, and Judith Tonhauser.
\newblock {The CommitmentBank}: Investigating projection in naturally occurring discourse.
\newblock 2019.
\newblock To appear in proceedings of Sinn und Bedeutung 23. Data can be found at https://github.com/mcdm/CommitmentBank/.

\bibitem{vit}
Alexey Dosovitskiy, Lucas Beyer, Alexander Kolesnikov, Dirk Weissenborn, Xiaohua Zhai, Thomas Unterthiner, Mostafa Dehghani, Matthias Minderer, Georg Heigold, Sylvain Gelly, Jakob Uszkoreit, and Neil Houlsby.
\newblock An image is worth 16x16 words: Transformers for image recognition at scale, 2021.

\bibitem{tiled_gemm}
Kayvon Fatahalian, Jeremy Sugerman, and Pat Hanrahan.
\newblock Understanding the efficiency of gpu algorithms for matrix-matrix multiplication.
\newblock In {\em Proceedings of the ACM SIGGRAPH/EUROGRAPHICS conference on Graphics hardware}, pages 133--137, 2004.

\bibitem{switch_transformers}
William Fedus, Barret Zoph, and Noam Shazeer.
\newblock Switch transformers: Scaling to trillion parameter models with simple and efficient sparsity, 2022.

\bibitem{giampiccolo2007third}
Danilo Giampiccolo, Bernardo Magnini, Ido Dagan, and Bill Dolan.
\newblock The third {PASCAL} recognizing textual entailment challenge.
\newblock In {\em Proceedings of the ACL-PASCAL workshop on textual entailment and paraphrasing}, pages 1--9. Association for Computational Linguistics, 2007.

\bibitem{openwebtext}
Aaron Gokaslan, Vanya Cohen, Ellie Pavlick, and Stefanie Tellex.
\newblock Openwebtext corpus.
\newblock \url{http://Skylion007.github.io/OpenWebTextCorpus}, 2019.

\bibitem{mamba}
Albert Gu and Tri Dao.
\newblock Mamba: Linear-time sequence modeling with selective state spaces, 2023.

\bibitem{huang2023moe}
Haiyang Huang, Newsha Ardalani, Anna Sun, Liu Ke, Hsien-Hsin~S. Lee, Anjali Sridhar, Shruti Bhosale, Carole-Jean Wu, and Benjamin Lee.
\newblock Towards moe deployment: Mitigating inefficiencies in mixture-of-expert (moe) inference, 2023.

\bibitem{inan_tying}
Hakan Inan, Khashayar Khosravi, and Richard Socher.
\newblock Tying word vectors and word classifiers: {A} loss framework for language modeling.
\newblock In {\em 5th International Conference on Learning Representations, {ICLR} 2017, Toulon, France, April 24-26, 2017, Conference Track Proceedings}. OpenReview.net, 2017.

\bibitem{applications_of_lms}
Jean Kaddour, Joshua Harris, Maximilian Mozes, Herbie Bradley, Roberta Raileanu, and Robert McHardy.
\newblock Challenges and applications of large language models, 2023.

\bibitem{nanogpt}
Andrej Karpathy.
\newblock nanogpt.
\newblock \url{https://github.com/karpathy/nanoGPT}, 2022.

\bibitem{khashabi2018looking}
Daniel Khashabi, Snigdha Chaturvedi, Michael Roth, Shyam Upadhyay, and Dan Roth.
\newblock Looking beyond the surface: A challenge set for reading comprehension over multiple sentences.
\newblock In {\em Proceedings of the 2018 Conference of the North American Chapter of the Association for Computational Linguistics: Human Language Technologies, Volume 1 (Long Papers)}, pages 252--262, 2018.

\bibitem{kitaev2020reformer}
Nikita Kitaev, Łukasz Kaiser, and Anselm Levskaya.
\newblock Reformer: The efficient transformer, 2020.

\bibitem{paged_attention}
Woosuk Kwon, Zhuohan Li, Siyuan Zhuang, Ying Sheng, Lianmin Zheng, Cody~Hao Yu, Joseph~E. Gonzalez, Hao Zhang, and Ion Stoica.
\newblock Efficient memory management for large language model serving with pagedattention, 2023.

\bibitem{levesque2011winograd}
Hector~J Levesque, Ernest Davis, and Leora Morgenstern.
\newblock The {W}inograd schema challenge.
\newblock In {\em {AAAI} Spring Symposium: Logical Formalizations of Commonsense Reasoning}, volume~46, page~47, 2011.

\bibitem{speculative_decoding}
Yaniv Leviathan, Matan Kalman, and Yossi Matias.
\newblock Fast inference from transformers via speculative decoding, 2023.

\bibitem{blockwise_parallel}
Hao Liu and Pieter Abbeel.
\newblock Blockwise parallel transformers for large context models.
\newblock In A.~Oh, T.~Naumann, A.~Globerson, K.~Saenko, M.~Hardt, and S.~Levine, editors, {\em Advances in Neural Information Processing Systems}, volume~36, pages 8828--8844. Curran Associates, Inc., 2023.

\bibitem{ring_attention}
Hao Liu, Matei Zaharia, and Pieter Abbeel.
\newblock Ring attention with blockwise transformers for near-infinite context, 2023.

\bibitem{splitwise}
Pratyush Patel, Esha Choukse, Chaojie Zhang, Íñigo Goiri, Aashaka Shah, Saeed Maleki, and Ricardo Bianchini.
\newblock Splitwise: Efficient generative llm inference using phase splitting, 2024.

\bibitem{formal_algorithms}
Mary Phuong and Marcus Hutter.
\newblock Formal algorithms for transformers, 2022.

\bibitem{pilehvar2018wic}
Mohammad~Taher Pilehvar and Jose Camacho-Collados.
\newblock {WiC}: The word-in-context dataset for evaluating context-sensitive meaning representations.
\newblock In {\em Proceedings of NAACL-HLT}, 2019.

\bibitem{poliak2018dnc}
Adam Poliak, Aparajita Haldar, Rachel Rudinger, J.~Edward Hu, Ellie Pavlick, Aaron~Steven White, and Benjamin {Van Durme}.
\newblock Collecting diverse natural language inference problems for sentence representation evaluation.
\newblock In {\em Proceedings of EMNLP}, 2018.

\bibitem{pope2023efficiently}
Reiner Pope, Sholto Douglas, Aakanksha Chowdhery, Jacob Devlin, James Bradbury, Jonathan Heek, Kefan Xiao, Shivani Agrawal, and Jeff Dean.
\newblock Efficiently scaling transformer inference.
\newblock {\em Proceedings of Machine Learning and Systems}, 5, 2023.

\bibitem{press2016_tying}
Ofir Press and Lior Wolf.
\newblock Using the output embedding to improve language models.
\newblock In {\em Conference of the European Chapter of the Association for Computational Linguistics}, 2016.

\bibitem{gpt2}
Alec Radford, Jeffrey Wu, Rewon Child, David Luan, Dario Amodei, and Ilya Sutskever.
\newblock Language models are unsupervised multitask learners.
\newblock {\em OpenAI blog}, 1(8):9, 2019.

\bibitem{roemmele2011choice}
Melissa Roemmele, Cosmin~Adrian Bejan, and Andrew~S. Gordon.
\newblock Choice of plausible alternatives: An evaluation of commonsense causal reasoning.
\newblock In {\em 2011 AAAI Spring Symposium Series}, 2011.

\bibitem{rudinger2018winogender}
Rachel Rudinger, Jason Naradowsky, Brian Leonard, and Benjamin {Van Durme}.
\newblock Gender bias in coreference resolution.
\newblock In {\em Proceedings of NAACL-HLT}, 2018.

\bibitem{multi_query}
Noam Shazeer.
\newblock Fast transformer decoding: One write-head is all you need, 2019.

\bibitem{shazeer2017outrageously}
Noam Shazeer, Azalia Mirhoseini, Krzysztof Maziarz, Andy Davis, Quoc Le, Geoffrey Hinton, and Jeff Dean.
\newblock Outrageously large neural networks: The sparsely-gated mixture-of-experts layer, 2017.

\bibitem{megatron_lm}
Mohammad Shoeybi, Mostofa Patwary, Raul Puri, Patrick LeGresley, Jared Casper, and Bryan Catanzaro.
\newblock Megatron-lm: Training multi-billion parameter language models using model parallelism, 2020.

\bibitem{original_attention}
Ashish Vaswani, Noam Shazeer, Niki Parmar, Jakob Uszkoreit, Llion Jones, Aidan~N. Gomez, Lukasz Kaiser, and Illia Polosukhin.
\newblock Attention is all you need, 2023.

\bibitem{super_glue}
Alex Wang, Yada Pruksachatkun, Nikita Nangia, Amanpreet Singh, Julian Michael, Felix Hill, Omer Levy, and Samuel~R. Bowman.
\newblock Superglue: {A} stickier benchmark for general-purpose language understanding systems.
\newblock {\em CoRR}, abs/1905.00537, 2019.

\bibitem{gpt-j}
Ben Wang and Aran Komatsuzaki.
\newblock {GPT-J-6B: A 6 Billion Parameter Autoregressive Language Model}.
\newblock \url{https://github.com/kingoflolz/mesh-transformer-jax}, May 2021.

\bibitem{overlap_comp_asplos2023}
Shibo Wang, Jinliang Wei, Amit Sabne, Andy Davis, Berkin Ilbeyi, Blake Hechtman, Dehao Chen, Karthik~Srinivasa Murthy, Marcello Maggioni, Qiao Zhang, Sameer Kumar, Tongfei Guo, Yuanzhong Xu, and Zongwei Zhou.
\newblock Overlap communication with dependent computation via decomposition in large deep learning models.
\newblock In {\em Proceedings of the 28th ACM International Conference on Architectural Support for Programming Languages and Operating Systems, Volume 1}, ASPLOS 2023, page 93–106, New York, NY, USA, 2022. Association for Computing Machinery.

\bibitem{orca_serving_system}
Gyeong-In Yu, Joo~Seong Jeong, Geon-Woo Kim, Soojeong Kim, and Byung-Gon Chun.
\newblock Orca: A distributed serving system for $\{$Transformer-Based$\}$ generative models.
\newblock In {\em 16th USENIX Symposium on Operating Systems Design and Implementation (OSDI 22)}, pages 521--538, 2022.

\bibitem{llmcompass}
Hengrui Zhang, August Ning, Rohan~Baskar Prabhakar, and David Wentzlaff.
\newblock Llmcompass: Enabling efficient hardware design for large language model inference.
\newblock In {\em To appear in Proceedings of the 51st Annual International Symposium on Computer Architecture}, 2024.

\bibitem{zhang2018record}
Sheng Zhang, Xiaodong Liu, Jingjing Liu, Jianfeng Gao, Kevin Duh, and Benjamin~Van Durme.
\newblock {ReCoRD}: Bridging the gap between human and machine commonsense reading comprehension.
\newblock {\em arXiv preprint 1810.12885}, 2018.

\end{thebibliography}

\appendix

\newpage

\section{Appendix}

\subsection{\textbf{Training setup and compute requirements}}
\label{appendix:training_setup_compute}

\begin{table}
\centering
\caption{\textbf{Pretraining compute and setup for each \archname~configuration}}
\resizebox{\textwidth}{!}{
\begin{tabular}{ P{0.225\textwidth}cP{0.125\textwidth}P{0.1\textwidth}P{0.1\textwidth}P{0.125\textwidth}P{0.125\textwidth}}
\toprule
\textbf{Model} & \textbf{Layers} & \textbf{Embedding Dimension} &  \textbf{Attention Heads} & \textbf{Total Params.} & \textbf{A100 GPU Hrs} & \textbf{Initial Learning Rate} \\
\toprule
  \archname\ 2-way & 12 & 678 & 12 & 124M & 480 & $2.5e-4$  \\
 \archname\ 4-way & 12 & 504 & 12 & 124.5M & 500 & $2.5e-4$\\
  \archname\ 6-way & 12 & 418 & 12 & 123.2M & 800 & $2.5e-4$ \\
\midrule
  \archname\ 2-way &  24 & 888 & 16 & 350M & 1000 & $1.5e-4$ \\
  \archname\ 4-way &  24 & 644 & 16 & 353.4M & 1750 & $1.5e-4$ \\
 \midrule
 \archname\ 4-way & 24 & 960 & 16 & 761M & 1800 & $1.5e-4$ \\
\bottomrule
\end{tabular}}
\label{tab:pretraining_details}
\end{table}

For all pretrained models presented in Section~\ref{sub_sec:model_confs_perplexity}, we used a similarly sized GPT-3~\cite{lms_fewshot_learners} model's hyperparameters as the basis and followed the procedure outlined in Section~\ref{sub_sub_sec:deriving_model_configs} to calculate the embedding dimension.
We did not make an effort to optimize the codebase used for training which builds off of nanoGPT~\cite{nanogpt}.
It is possible to replicate pretrained models by extending nanoGPT to implement the new forward pass as described in Algorithm~\ref{alg:forward_pass}.
The Adam optimizer was used to train all models along with a cosine learning rate decay with linear warmup.
Initial learning rates and the approximate GPU hours required to train each configuration are presented in Table~\ref{tab:pretraining_details}.
All models were trained for $300,000$ gradient steps.
Only the $761M$ parameter model was trained on a node with 80GB A100 GPU machines.
The other configurations were trained on 40GB A100 machines and consequently use many more gradient accumulation steps.
This is why the largest model required a similar number of GPU hours as the next largest.

Weights for all pretrained GPT-2 models used when evaluating SuperGLUE performance~\cite{demarneffe:cb}~\cite{roemmele2011choice}~\cite{khashabi2018looking}~\cite{zhang2018record}~\cite{dagan2006pascal}~\cite{bar2006second}~\cite{giampiccolo2007third}~\cite{bentivogli2009fifth}~\cite{pilehvar2018wic}~\cite{rudinger2018winogender}~\cite{poliak2018dnc}~\cite{levesque2011winograd} in Section~\ref{sub_sec:superglue_perf} were obtained from HuggingFace.
Since the focus of this work is on illustrating the efficiency gains and language modeling capabilities, we do not implement any safeguards that will filter for biased and/or harmful content.
Initial experiments that tried various variations of the model architecture required about another $1,000$ hours of A100 compute.

\subsection{Overlap plugin implementation}
\label{appendix:overlap_plugin_implementation}

We used TensorRT-LLM version \textit{0.12.0.dev2024073000} throughout the evaluation.
CUDA allows kernels to be launched on different streams and depending on resource availability, these kernels may be executed in parallel.
However, TensorRT does not support multi-stream execution across plugins.
We circumvented this limitation by implementing a Singleton that can manage a dedicated stream and global memory meant for launching collectives.
This allows different instances of the plugin to launch and synchronize kernels on the same stream.
Each instance of the plugin can either: trigger an AllReduce op on a separate low-priority CUDA stream or synchronize the stream to ensure that a previously launched AllReduce completes.
The plugin also implements the functionality provided in the existing GEMM plugin.
This allows us to perform the following within the Multi-Head Attention block:
\begin{enumerate}
    \item  The AllReduce is launched on a dedicated CUDA stream just before the GEMM used to compute the Query, Key, and Value matrices (GEMM $ Q\_K\_V$~\ref{fig:transformer_layer}) via the plugin
    \item All intermediate compute is performed using existing kernels such as FlashAttention
    \item The CUDA stream that the AllReduce op was placed on is synchronized after the GEMM $Wo\_proj$~\ref{fig:transformer_layer}, also via the Plugin
\end{enumerate}
This approach allowed us to overlap the collective with all the computation required for Multi-Head Attention and is the only addition to TensorRT-LLM aside from the definition of \archname.
However, it also requires two extraneous memory copy operations that can be avoided if the library adds support for multi-stream execution.

\subsection{Runtime characterization}
\label{appendix:runtime_characterization}

\begin{figure}[ht]
\centering
\includegraphics[width=1.0\textwidth]{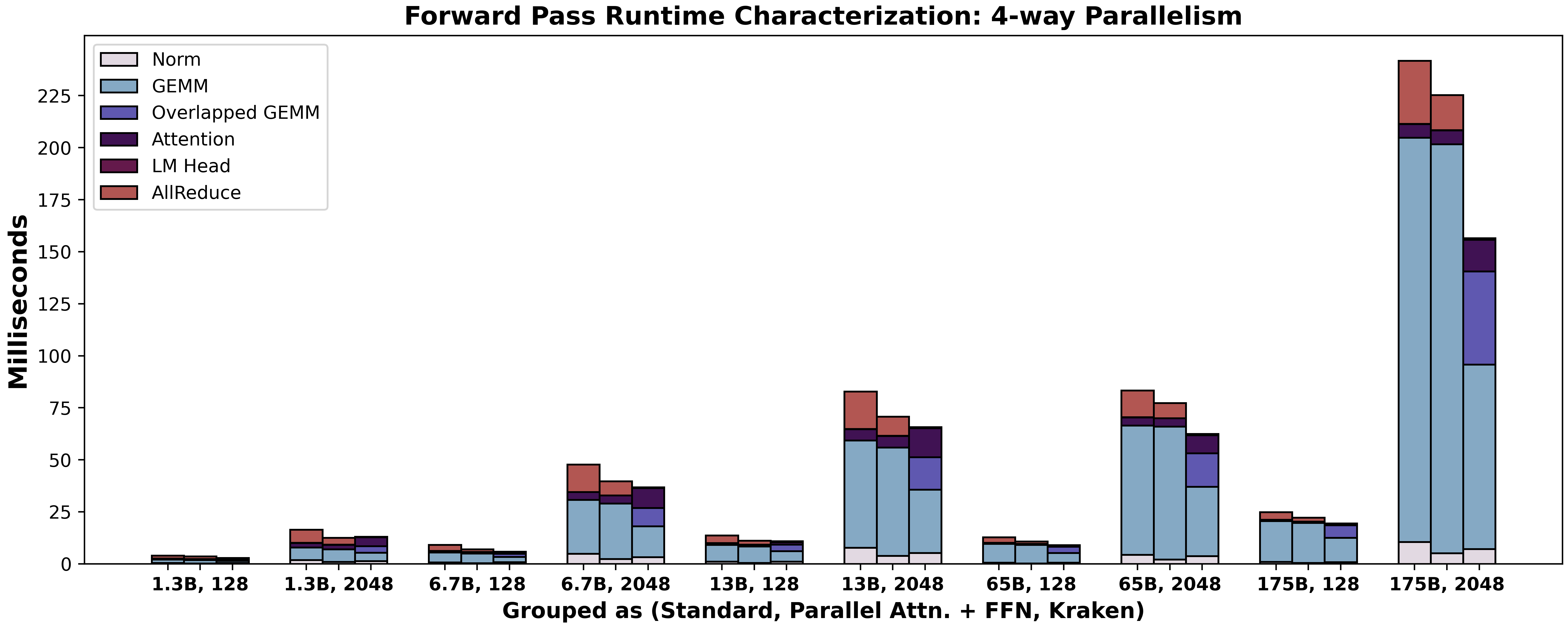}
\caption{\textbf{Runtime Characterization: 4-way Parallelism.} For each cluster on the x-axis, labels denote the size of model followed by the context length.}
\label{fig:profile_4}
\vspace{-0.2cm}
\end{figure}

\begin{figure}[ht]
\centering
\includegraphics[width=1.0\textwidth]{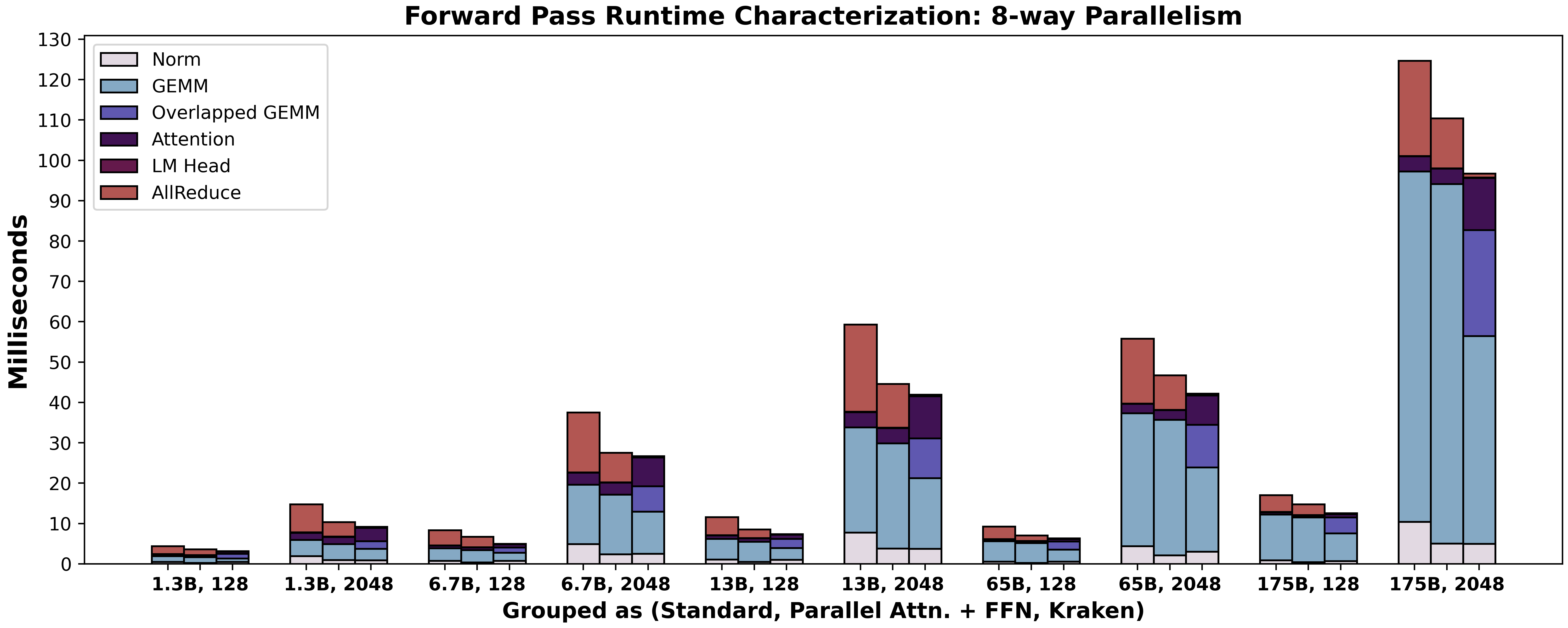}
\caption{\textbf{Runtime Characterization: 8-way Parallelism.} For each cluster on the x-axis, labels denote the size of model followed by the context length.}
\label{fig:profile_8}
\vspace{-0.2cm}
\end{figure}

To provide more context to the performance gains presented in Figures~\ref{fig:latency_improvement_4} and~\ref{fig:latency_improvement_8}, we characterize the entire forward pass using the same experimental setting as Section~\ref{sub_sec:first_token_latency}.
The runtime profiles presented in Figure~\ref{fig:profile_4} and Figure~\ref{fig:profile_8} were obtained using the profiler built into \textit{gptSessionBenchmark}.
Across all model sizes and both context lengths, we find that a significant proportion of time is spent in the \texttt{AllReduce}.
This proportion increases for 8-way parallelism compared to 4-way parallelism which is expected since each device works on a smaller fraction of the compute.
In some cases such as for 175B models at context length 2048, some of the performance gains come from the GEMMs requiring significantly more time for the GPT and GPT-J like configurations.
Despite these outliers, in general, \archname~models spend a much smaller fraction of runtime in inter-device communication leading to considerable performance and efficiency improvements.
Note that the cost of the memory copies and synchronization necessary for the Overlap plugin are implicitly included in the \texttt{Overlapped GEMM} fraction.
More precisely, \texttt{Overlapped GEMM} is the time spent computing GEMM $Wo\_proj$, GEMM $Q\_K\_V$, and other operations in the Overlap plugin.

\subsection{\textbf{Time To First Token in milliseconds}}
\label{appendix:ttft_seconds}

\begin{table}
\centering
\caption{\textbf{Inference latency in milliseconds for 4-way parallelism}}
\label{tab:latency_ms_4}

\begin{tabular}{ccccc}

\toprule

\textbf{Model Size} & \textbf{Context Length} & \textbf{Standard} & \textbf{Parallel Attn. + FeedForward} & \textbf{Kraken} \\
\toprule

1.3B &  128 & 3.7 & 3.3 & 3.3 \\
1.3B &  2048 & 17.0 & 13.6 & 13.2 \\
\midrule
6.7B &  128 & 8.3 & 7.0 & 5.8 \\
6.7B &  2048 & 48.2 & 42.1 & 38.0 \\
\midrule
13B &  128 & 13.0 & 11.1 & 10.7 \\
13B &  2048 & 83.7 & 73.8 & 66.9 \\
\midrule
65B &  128 & 12.7 & 11.2 & 8.5 \\
65B &  2048 & 84.6 & 79.5 & 63.8 \\
\midrule
175B &  128 & 24.6 & 22.5 & 19.9 \\
175B &  2048 & 243.7 & 230.8 & 158.9 \\
\bottomrule

\end{tabular}
\end{table}

\begin{table}
\centering
\caption{\textbf{Inference latency in milliseconds for 8-way parallelism}}
\label{tab:latency_ms_8}

\begin{tabular}{ccccc}

\toprule

\textbf{Model Size} & \textbf{Context Length} & \textbf{Standard} & \textbf{Parallel Attn. + FeedForward} & \textbf{Kraken} \\
\toprule

1.3B &  128 & 4.3 & 3.4 & 3.2 \\
1.3B &  2048 & 15.9 & 11.7 & 9.7 \\
\midrule
6.7B &  128 & 7.1 & 5.7 & 4.7 \\
6.7B &  2048 & 37.3 & 29.2 & 27.6 \\
\midrule
13B &  128 & 10.7 & 8.4 & 6.7 \\
13B &  2048 & 58.8 & 46.9 & 42.6 \\
\midrule
65B &  128 & 8.7 & 7.2 & 6.2 \\
65B &  2048 & 55.9 & 48.9 & 42.9 \\
\midrule
175B &  128 & 16.9 & 14.4 & 12.4 \\
175B &  2048 & 125.1 & 114.7 & 98.0 \\
\bottomrule

\end{tabular}
\end{table}

Table~\ref{tab:latency_ms_4} contains the results from Figure~\ref{fig:latency_improvement_4} but in terms of milliseconds.
Similarly, Table~\ref{tab:latency_ms_8} presents the results from Figure~\ref{fig:latency_improvement_8}.
Note that the 65B and 175B configurations are scaled to about $\frac{1}{4}$ the actual runtime as their engines are built with only a quarter of the layers due to device memory capacity constraints.

\subsection{\textbf{Effect on KV Cache size}}
\label{appendix:kv_cache_analysis}

\begin{figure}[!h]
\centering
\begin{minipage}[c]{0.48\linewidth}
\centering
\includegraphics[scale=0.46]{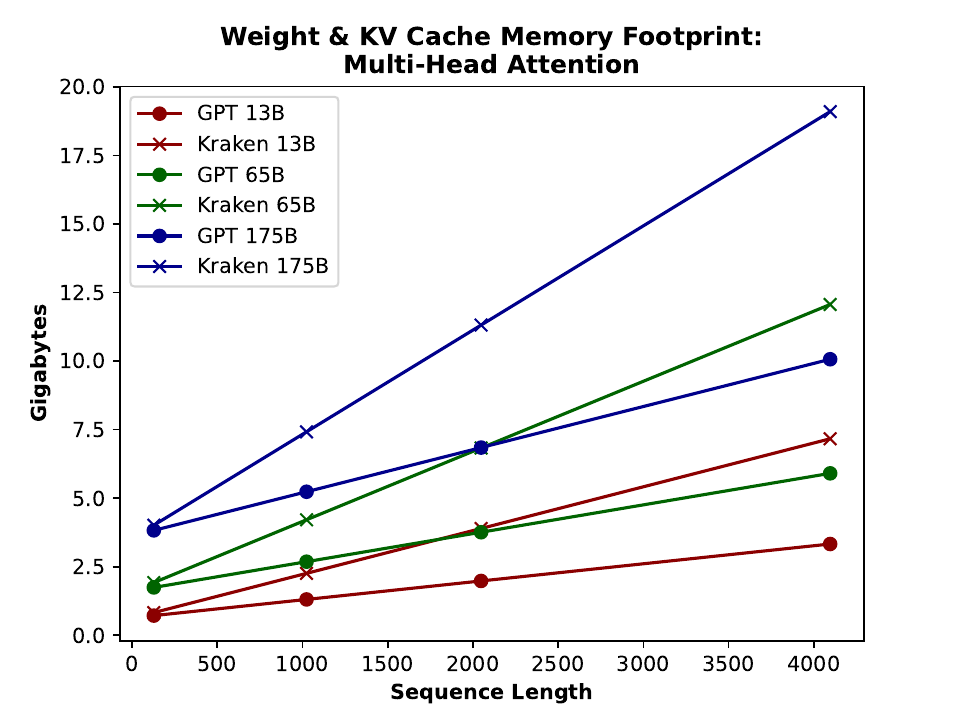}
\caption{\textbf{The memory footprint of each \archname~layer compared to that of a similar standard Transformer layer at batch size 32 when Multi-Head Attention is used.}}
\label{fig:kv_cache_multi_head}
\end{minipage} \hfill
\begin{minipage}[c]{0.48\linewidth}
\centering
\includegraphics[scale=0.46]{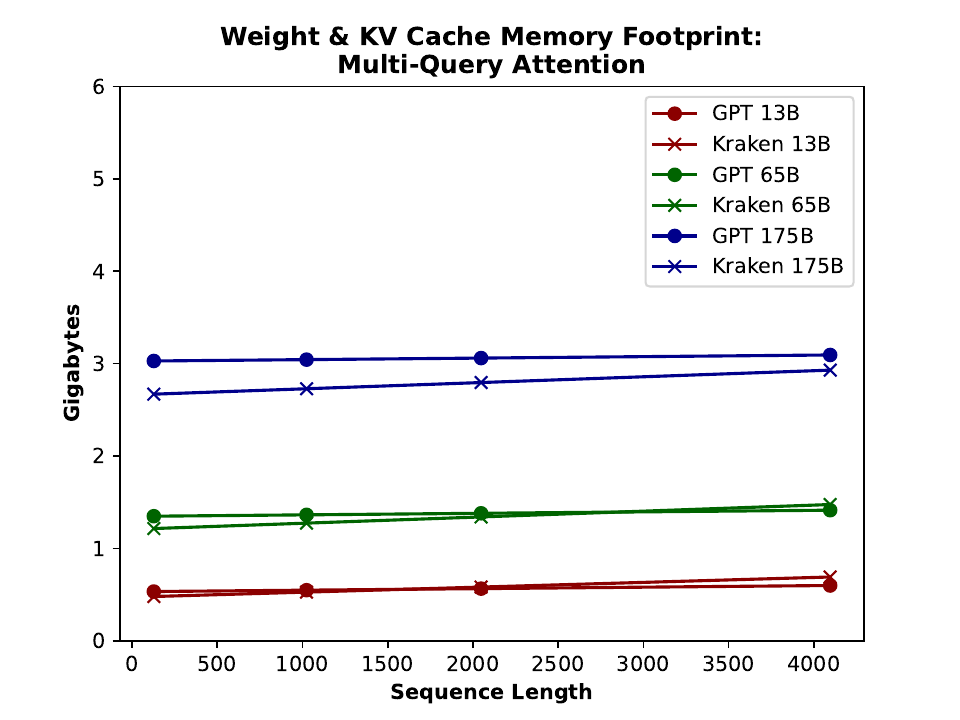}
\caption{\textbf{Using Multi-Query Attention greatly reduces the increased memory consumption that is incurred by \archname~layers.}}
\label{fig:kv_cache_multi_query}
\end{minipage}
\end{figure}

A consequence of following the procedure detailed in Section~\ref{sub_sub_sec:deriving_model_configs} is that while the embedding dimension is smaller, the absolute size of each layer's KV Cache will be larger.
This effect is depicted in Figure~\ref{fig:kv_cache_multi_head} which compares \archname~layers with similar standard Transformer layers.
Here, we assume that weights and the cache are stored in half-precision (16-bit) and calculate the total memory footprint of a single layer for the three largest configurations detailed in Table~\ref{tab:engine_model_dims}.
Batch size was fixed at a relatively high $32$ but we do not include any memory required by activations.
Given that this effect can become noticeable at long sequence lengths, one strategy to mitigate it is by adopting either Multi-Query Attention~\cite{multi_query} or Grouped-Query Attention~\cite{ainslie2023gqa} as demonstrated in Figure~\ref{fig:kv_cache_multi_query}.
Note that the layers in the Multi-Query setting are smaller because the linear transformation used to compute $QKV$ is smaller since \textit{Keys} and $\textit{Values}$ are shared across heads.

\end{document}